\pdfoutput=1

\documentclass[11pt]{article}

\usepackage{coling}
\usepackage{arydshln}
\usepackage{times}
\usepackage{latexsym}

\usepackage[T1]{fontenc}

\usepackage[utf8]{inputenc}

\usepackage{microtype}

\usepackage{inconsolata}

\usepackage{graphicx}

\usepackage{tcolorbox}

\usepackage{xcolor}

\usepackage{enumitem}

\usepackage{listings}
\usepackage{placeins}
\usepackage{float}
\usepackage{afterpage}

\usepackage{stfloats}

\def\code#1{\texttt{#1}}

\title{\textsc{Karrierewege}: A Large Scale Career Path Prediction Dataset}

\author{Elena Senger$^{1,2}$ \quad Yuri Campbell$^{2}$ \quad Rob van der Goot$^{3}$ \quad Barbara Plank$^{1}$ \\[1.5ex]
$^1$MaiNLP, Center for Information and Language Processing, LMU Munich, Germany \\
$^2$Fraunhofer Center for International Management and Knowledge Economy IMW, Germany \\
$^3$Department of Computer Science, IT University of Copenhagen, Denmark \\
\texttt{elena.senger@cis.lmu.de, yuri.campbell@imw.fraunhofer.de} \\
\texttt{robv@itu.dk, b.plank@lmu.de}
}

\begin{document}
\maketitle
\begin{abstract}
Accurate career path prediction can support many stakeholders, like job seekers, recruiters, HR, and project managers. However, publicly available data and tools for career path prediction are scarce.
In this work, we introduce \textsc{Karrierewege}, a comprehensive, publicly available dataset containing over 500k career paths, significantly surpassing the size of previously available datasets. We link the dataset 
to the ESCO taxonomy to offer a valuable resource for predicting career trajectories. To tackle the problem 
of free-text inputs typically found in resumes, we 
enhance it by synthesizing 
job titles and descriptions resulting in \textsc{Karrierewege+}. This allows for accurate predictions from unstructured data, closely aligning with real-world application challenges. We benchmark existing state-of-the-art (SOTA) models on our dataset and a prior benchmark and  observe improved performance and robustness, particularly for free-text use cases, due to the synthesized data.
\end{abstract}

\section{Introduction}
Career path prediction (also known as career trajectory prediction) is a growing field \citep{Shreyas.2024}, with the potential to inform recruitment, career counseling, upskilling or reskilling, or more broadly workforce planning and workforce trends. The task is to predict future career moves based on an individual's work history, possibly using further information such as skills or education. To achieve this, robust datasets that capture detailed career histories are essential. However, the availability of large-scale benchmark career history datasets remains limited \citep{du2024laborllmlanguagebasedoccupationalrepresentations, decorte.2023}, posing a major challenge for the field. 

A dataset mapped to ESCO (European Skills, Competences, Qualifications, and Occupations) Taxonomy is particularly advantageous because ESCO provides a standardized ``common language'' for occupations and skills across the European labor market, describing over 3,000 occupations and nearly 14,000 skills in 28 languages.\footnote{\url{https://esco.ec.europa.eu/en/about-esco/what-esco}} 
Since its introduction in 2017, ESCO has attracted diverse stakeholders—including employment services, job portals, educational institutions, HR departments, and international organizations.\footnote{\url{https://esco.ec.europa.eu/en/about-esco/esco-stakeholders}} 

Building on these insights, we release a large, publicly available dataset mapped to ESCO occupations to address the critical need for comprehensive and standardized resources in career path prediction. By leveraging ESCO's ``common language'' for occupations and skills, our dataset aims to foster research and development in this growing field, paving the way for more accurate career trajectory modeling. We document the steps involved in our dataset creation process to encourage the development and evaluation of customized datasets tailored to real-world applications,
ultimately promoting job mobility and fostering a more integrated and efficient labor market.
Our contributions are:
\begin{itemize}[itemsep=1pt, topsep=1pt, parsep=1pt, partopsep=1pt]
    \item Introducing \textsc{Karrierewege}, a new large-scale career path prediction dataset consisting of over 500,000 career paths.
    \footnote{Karrierewege: \url{https://huggingface.co/datasets/ElenaSenger/Karrierewege}\\ Karrierewege plus: \url{https://huggingface.co/datasets/ElenaSenger/Karrierewege_plus}} 
    \item Mapping the dataset to the ESCO taxonomy, enhancing interoperability and facilitating research and real-world applications that utilize ESCO. 
    \item Exploring data synthesis techniques by generating paraphrased titles and descriptions from the taxonomical dataset to address the real-world challenge of free text data. 
    \item A reproduction study of \citet{decorte.2023} to compare results on their benchmark dataset with the newly introduced \textsc{Karrierewege} and \textsc{Karrierewege+} datasets. 
\end{itemize}

\section{Related Work}
\subsection{Datasets for Career Path Prediction}

In the literature on machine learning-based career path prediction, most prior work uses large \textit{non-public} datasets, typically sourced from major career portals such as LinkedIn \citep{Li.2027, CerillaSVF23}, Randstad \citep{schellingerhout2022explainable}, or Zippia \citep{vafa2024careerfoundationmodellabor}. As publicly available datasets, survey data is a popular choice \citep{Chang2019ABA, vafa2024careerfoundationmodellabor, du2024laborllmlanguagebasedoccupationalrepresentations} -- see Table \ref{tab:datasets}. But survey data is typically relatively small, or does not track the same individuals over a longer time span. For example, the Current Population Survey---a national U.S. labor force survey used in \citet{Chang2019ABA}---has a panel of 54,000 respondents per year but contains a person's occupation for only two consecutive years. Other surveys  \cite{vafa2024careerfoundationmodellabor,du2024laborllmlanguagebasedoccupationalrepresentations} are relatively small with sizes around 9-12k respondents. A small publicly available dataset is introduced by \citet{decorte.2023}. It is created using a Kaggle dataset of 2,482 anonymized English resumes. All occupations are linked to ESCO (version 1.1.2). The dataset includes both self-written job titles and synthetic descriptions (grounded on resumes), as well as standardized ESCO titles. Inspired by~\citet{decorte.2023}, 
we use their SOTA approach 
and compare results to their smaller dataset (see further Section~\ref{sec:datasetcomp}).
\begin{table}[h!]
  \centering
  \resizebox{\columnwidth}{!}{
  \begin{tabular}{p{4.7cm}|p{5.3cm}|r}
    \hline
    \textbf{Dataset} & \textbf{Paper} & \textbf{Size} \\
    \hline
    Nat. Longitudinal Survey of Youth 1979 & \citet{vafa2024careerfoundationmodellabor},   \citet{du2024laborllmlanguagebasedoccupationalrepresentations} & 1,200 \\
    Nat. Longitudinal Survey of Youth 1997 & \citet{vafa2024careerfoundationmodellabor}, \citet{du2024laborllmlanguagebasedoccupationalrepresentations} & 9,000 \\
    Panel Study of Income Dynamics & \citet{vafa2024careerfoundationmodellabor},  \citet{du2024laborllmlanguagebasedoccupationalrepresentations} & 12,000 \\
    Current Population Survey* & \citet{Chang2019ABA} & 54,000 \\
    DECORTE & \citet{decorte.2023} & 2,000 \\
    \textsc{Karrierewege} & our paper & 500,000 \\
    \textsc{Karrierewege+} & our paper & 100,000 \\
    \hline
  \end{tabular}
  }
  \caption{\label{tab:datasets} Summary of datasets used in various studies. *Only data for two consecutive years per person.}
\end{table}


\subsection{Methods for Synthetic Data Generation and LLMs in Occupations}
We source the original raw data from the German Employment Agency, but it only includes standardized job titles and descriptions. To make the model more applicable to real-world scenarios, where resumes often use varied, paraphrased job titles, we generated synthetic training data. This allows for more robust career path prediction models that can handle the complexities of free-text inputs.

Off-the-shelf (non-fine-tuned) large language models (LLMs) have been successfully applied to paraphrasing tasks across various domains \citep{Jayawardena_2024} and have also been used to generate synthetic data in the job market, such as to create job vacancies \citep{li2023llm4jobsunsupervisedoccupationextraction, magron-etal-2024-jobskape}. Their effectiveness in representing occupations likely stems from the extensive training of LLMs on diverse sources of data, including occupational data,  labor market news and job-related texts \citep{du2024laborllmlanguagebasedoccupationalrepresentations}. This demonstrated success in the occupational domain supports our approach of leveraging LLMs for synthesizing training data by paraphrasing job titles and generating corresponding descriptions.



\section{\textsc{Karrierewege}}

\begin{table*}
  \centering
  \resizebox{2\columnwidth}{!}{
  \begin{tabular}{c|c|p{4.1cm}|p{4cm}|p{4cm}|p{4cm}|p{4cm}|p{4cm}}
    \hline
    \textbf{id} & \textbf{order} & \textbf{ESCO\_title} & \textbf{ESCO\_description} & \textbf{new\_title\_oc} & \textbf{new\_description\_oc} & \textbf{new\_title\_cp} & \textbf{new\_description\_cp} \\
    \hline
    0 & 0 & Medical laboratory manager & Medical laboratory managers oversee ... & Quality Assurance Specialist & The Quality Assurance Specialist is ... & Laboratory Director & Laboratory Director: Oversees ... \\
    \hline
    0 & 1 & Environmental health inspector & Environmental health inspectors carry ... & Pollution Control Specialist & The Pollution Control Specialist is responsible ... & Environmental Safety Specialist & Environmental Safety Specialist: Conducts ... \\
    \hline
    0 & 2 & Environmental health inspector & Environmental health inspectors carry ... & Environmental Compliance Officer & Environmental Compliance Officer: Conducts ... & Environmental Safety Specialist & Environmental Safety Specialist: Conducts ... \\
    \hline
    1 & 0 & Food service worker & Food service workers prepare food and ... & Concession Stand Staff & Operate a concession stand, selling ... & Culinary Service Provider & Culinary Service Provider: Provides expert culinary ... \\
    \hline
    1 & 1 & Dietitian & Dietitians assess specific nutritional ... & Eating Disorder Specialist & The Eating Disorder Specialist provides ... & Nutritionist & Nutritionist: Helps people develop healthy ... \\
    \hline
  \end{tabular}
  }
  \caption{\label{tab:job_mapping_simplified} Example entries from the \textsc{Karrierewege+} validation dataset. Rows sharing the same \textit{id} refer to work experiences of the same person, with \textit{order} indicating the sequence. Titles and descriptions with the \textit{\_oc} suffix are synthesized per occupation, while those with \textit{\_cp} are synthesized per career path.}
\end{table*}

\begin{figure*}
  \centering
  \includegraphics[width=1.0\linewidth]{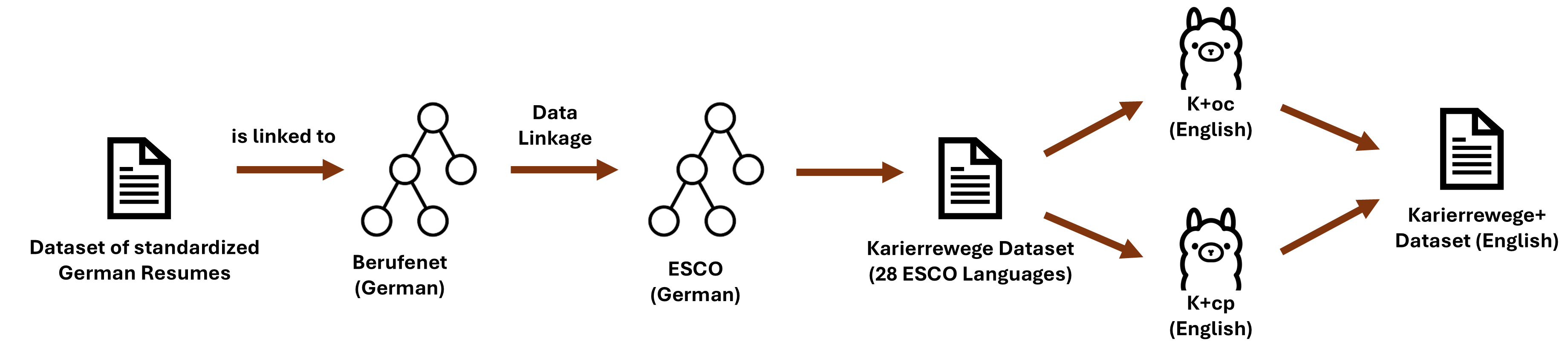}
  \caption{Steps necessary to create \textsc{Karrierewege} and \textsc{Karrierewege+} datasets. Titles and descriptions with the \textit{\_oc} suffix are synthesized per occupation, while those with \textit{\_cp} are synthesized per career path.}
  \label{fig:datasets}
\end{figure*}

To create a large and diverse dataset for career path prediction, we sourced the data from anonymized resumes provided by the German employment agency as a basis (see Figure \ref{fig:datasets} for the dataset creation process).\footnote{\url{https://www.arbeitsagentur.de/bewerberboerse/}} This dataset encompasses resumes from individuals seeking employment across all industries. We note that despite of its size, the resulting dataset may still be biased---it possibly contains more resumes from industries with lower demand (where individuals are more inclined to register as unemployed) than from high-demand industries, where unemployment registration is less common. Additionally, since all resumes are from individuals seeking employment in Germany, there is a cultural bias towards that region. 

\subsection{Mapping to ESCO}

 Due to restrictions preventing the direct publication of these anonymized CVs, and in recognition of the widespread adoption of the ESCO framework, we manually mapped occupations from the German resumes to their equivalents in the German ESCO taxonomy (version 1.2.0). This mapping ensures compatibility with the widely adopted ESCO framework, enriches the dataset with additional attributes like skills and job descriptions, and enables accessibility in 28 languages. For consistency with previous work, we use the English ESCO attributes in this paper. Yet, the published dataset can be converted to any of the other languages via the unique identifiers provided by ESCO.

The raw dataset contains standardized occupational titles from the German Berufenet taxonomy (version 2020).\footnote{\url{https://web.arbeitsagentur.de/berufenet/}} Both Berufenet and ESCO occupations are mapped to ISCO-08 codes. However, due to broad categorization inherent in ISCO-08, multiple occupations share the same code, making it unsuitable for direct one-to-one mapping.
Therefore, to link Berufenet and ESCO, we experimented with three methods as outlined next. 
 The first method involved embedding similarity, using the \code{distiluse-base-multilingual}  model to calculate semantic similarity based on job titles and descriptions. The second method utilized the ESCO API, which returned less accurate mappings due to incomplete queries and higher-level job titles. The third method involved GPT models, specifically \code{GPT-3.5} and \code{GPT-4o-mini} for ranking mappings.
Overall, \code{GPT-4o-mini} performed best, particularly when using English prompts, achieving around 60\% correct mappings (see Table \ref{tab:embedding_similarity}). However, ultimately, none of these approaches consistently produced satisfactory results. Hence, we used them only to speed up the manual linkage process conducted by a trained assistant and one of the authors.

\begin{table}
\resizebox{\columnwidth}{!}{
  \centering
  \begin{tabular}{cccc}

    \hline
    Method & \% Correct Links \\
    \hline
    Embedding Similarity Title & 51.1 \\
    Embedding Similarity Description & 51.2 \\
    ESCO API & 30.7 \\
    GPT 3.5 DE Prompt & 42.6 \\
    GPT 3.5 EN Prompt & 52.9 \\
    GPT 4o mini DE Prompt & 59.4 \\
    GPT 4o mini EN Prompt & 60.4 \\
    \hline
  \end{tabular}
    }
  \caption{\label{tab:embedding_similarity}
    Percentage of correct links per method for mapping ESCO and Berufenet occupations.
  }
\end{table}

\subsection{Filtering the Data}
We excluded all resumes with missing entries in the work history field and kept only those resumes with more than one and less than thirty work experiences. We also kept only resumes with a change of occupation in their career history, as we are particularly interested in learning and predicting these. Additionally, we excluded resumes that contained rare occupations (less than 10 times in the dataset). This resulted in 568,888 resumes. 

\section{\textsc{Karrierewege+}: Synthesized Data}

\subsection{Generating Free-Text Job Titles}

To generate free-text data, we use two data synthesizes methods:

\paragraph{\textsc{Karrierewege+}oc}

In the first approach, we use \code{LLAMA 3.1 8b} to generate seven alternative titles for each ESCO occupation title (K+oc). The choice of seven paraphrased titles was based on empirical observations indicating that generating more than seven titles often resulted in lower quality titles. For each paraphrased title, we additionally generated a corresponding job description using the same model. The underlying hypothesis for this method is that the paraphrased titles remain closely related to the original titles while being sufficiently distinct from other ESCO titles. 

\paragraph{\textsc{Karrierewege+}cp}

In the second approach, we directly synthesized the entire sequence of titles of a career path (K+cp). The hypothesis guiding this approach is that providing the model with the context of previous and subsequent occupation titles enables it to generate more appropriate and contextually relevant paraphrased titles. This method aims to achieve higher diversity by paraphrasing each ESCO occupation title more frequently, thereby introducing slight variations and increasing the richness of the dataset. Similar to the first approach, a corresponding job description was generated for each paraphrased title. The language model used was again the \code{LLAMA 3.1 8b} model. Due to the computational intensity of synthesizing individual career paths, we limited this approach to a random subset of 100,000 resumes. To maintain comparability between the two synthesis methods, we restricted the first synthesizing approach to the same number of resumes. All prompts are provided in Appendix \ref{sec:prompts-generation}.

\subsection{Evaluating the Quality of Paraphrased Titles and Career Paths}

\subsubsection{Quantitative Analysis}

To evaluate the quality of the paraphrased job titles and descriptions, we followed best practices recommended by \citet{van-der-lee-etal-2019-best} for paraphrase evaluation, i.e., to use well-defined evaluation criteria, avoid the use of smaller scales in rating (e.g., 2-point or 3-point Likert scale), employing a within-subjects design (where evaluators reviewed outputs from all systems),  randomized orderings to mitigate bias from order effects and complement subjective with objective measures to provide a comprehensive evaluation. Following  \citet{Jayawardena_2024}, we used BLEU \citep{papineni-etal-2002-bleu}, ROUGE-L \citep{lin-2004-rouge}, and a 5-point Likert scale to evaluate the quality of the paraphrased job titles. We overall assessed the generated paraphrases on four key dimensions:

\begin{itemize}[itemsep=0.5pt, topsep=1pt]
    \item \textbf{Correctness}: Measures if paraphrased titles are valid job titles and distinct from the original. Scores range from 0 (invalid or identical titles) to 5 (all titles valid and distinct).
    \item \textbf{Semantic Similarity}: Evaluates how well paraphrased titles capture the meaning of the original titles. Scores range from 0 (low similarity) to 5 (high similarity).
    \item \textbf{Diversity}: Assesses variety in the paraphrased career paths with a score of 0 indicating repetition, while 5 reflects a wide range of titles.
    \item \textbf{Coherence}: Measures the logical coherence of the paraphrased titles with the career path, where 0 means titles do not form a logical progression, and 5 indicates a coherent sequence.
\end{itemize}

To evaluate these dimensions, we manually labeled 100 resumes. One author, unaware of which synthesis method was used, manually evaluated each resume on the four dimensions. To further validate our findings, we used \code{GPT-4o mini} to evaluate the same metrics, after checking the alignment on the 100 manually labeled samples. We experimented with two prompt versions: one where the model was prompted once for all metrics, and another where the model was prompted for each metric individually (see the Appendix \ref{sec:prompts-evaluation} for the prompts). Following the best practice of \citet{thakur2024judgingjudgesevaluatingalignment}, we used Cohen’s kappa as a measure of alignment. Cohen’s kappa and the mean values for each metric indicated that prompting the model for all metrics at once resulted in closer alignment with human judgments. In general, Cohen’s kappa values were relatively low, particularly for coherence, but showed stronger alignment for less subjective metrics like diversity and correctness (see Appendix \ref{sec:appendix_alignement_scores}
for detailed scores).
The human and LLM scores revealed that K+cp achieved higher mean scores in correctness, semantic similarity, and coherence compared to K+oc. However, the K+oc outperformed K+cp in terms of diversity. These results are consistent with our expectations: the K+cp processes the entire career path, allowing for more coherent title generation, while the K+oc tends to produce greater diversity since it randomly selects from seven paraphrased options for each occupation. Overall, the Likert scale scores suggest that the K+cp yields higher-quality paraphrases.

As objective complementary measures, we used BLEU and ROUGE-L, comparing sequences of job titles and descriptions across entire career paths. For BLEU, we applied a smoothed score to account for low n-gram overlaps. In both metrics, K+cp consistently achieved slightly higher values, indicating better lexical similarity and sequence alignment with the original labels. However, the overall low scores suggest notable differences between both methods and the original career path (see Appendix \ref{sec:appendix_alignement_scores}
for detailed scores).

\subsubsection{Qualitative Analysis}
\begin{table*}
\centering
\resizebox{2\columnwidth}{!}{
\begin{tabular}{l|l|l}
\hline
\textbf{Category} & \textbf{Original Job Title} & \textbf{Paraphrased Job Title} \\ 
\hline
\textbf{1. Increasing Professionalism} 
                                       & Cashier & Retail Sales Associate \\
                                       & Mover & Professional Mover \\
                                       & Bricklayer & Building Specialist \\
\hline
\textbf{2. Inaccurate or Non-existent Titles} & Technical Communicator & User Experience \\
                                              & Financial Broker & Wealth Manager: Helping clients... \\
                                              & Bicycle Courier & On-the-Go Logistics Professional \\
                                              & Printed Circuit Board Assembler & PCB \\
\hline
\textbf{3. Semantic Mismatch} & City Councillor & Urban Planner \\
                              & House Sitter & Home Care Provider \\
                              & Food Production Operator & Production Line Worker \\
                              & Foster Care Support Worker & Support Worker \\
\hline
\textbf{4. Career Path as Story} & Hairdresser $\rightarrow$ Sales Account Manager $\rightarrow$ & Beauty Professional $\rightarrow$ Business Developer $\rightarrow$ \\
                                 & Commercial Sales Rep $\rightarrow$ Hairdresser & Sales Specialist $\rightarrow$ Salon Owner/Operator \\
\cline{2-3} 
                                 & Vehicle Cleaner $\rightarrow$ Factory Hand $\rightarrow$ & Construction Laborer $\rightarrow$ Flooring Installer $\rightarrow$ \\
                                 & Metal Sawing Machine Operator & Floor Covering Technician \\
\hline
\end{tabular}
}
\caption{\label{tab:qualitative} Examples of qualitative analysis of paraphrased job titles}
\end{table*}

The paraphrased job titles reveal several interesting and distinct trends and errors. One common trend is the enhancement of professionalism, where paraphrased titles elevate the perceived professionalism of the original job titles, like ``Bricklayer'' becomes ``Building Specialist''. Another issue is the introduction of inaccurate or non-existent job titles, such as ``On-the-Go Logistics Professional''
where the paraphrasing becomes overly creative and diverges from widely recognized titles. Semantic mismatches also occur, for instance, when “city councillor,” a political position, is paraphrased as “urban planner,” a technical role focused on infrastructure. Lastly, when using the K+cp, the paraphrasing often constructs cohesive career paths, showing clear progression and skills development over time. However, when paraphrasing individual occupations without considering the full career path, this cohesive narrative is lost. Examples of these patterns are presented in Table \ref{tab:qualitative}.

\section{Dataset Statistics and Comparison}\label{sec:datasetcomp}

To ensure the practical utility of \textsc{Karrierewege} for real-world career path prediction, we present key statistics on the dataset's characteristics—including the number of resumes, the distribution and diversity of ESCO occupations and industries, and statistics on synthesized job titles and descriptions that reflect the complexity of free-text inputs. These insights demonstrate the dataset's comprehensiveness and its suitability for advancing both academic research and industrial applications.

\begin{figure}
  \centering
  \includegraphics[width=1.0\linewidth]{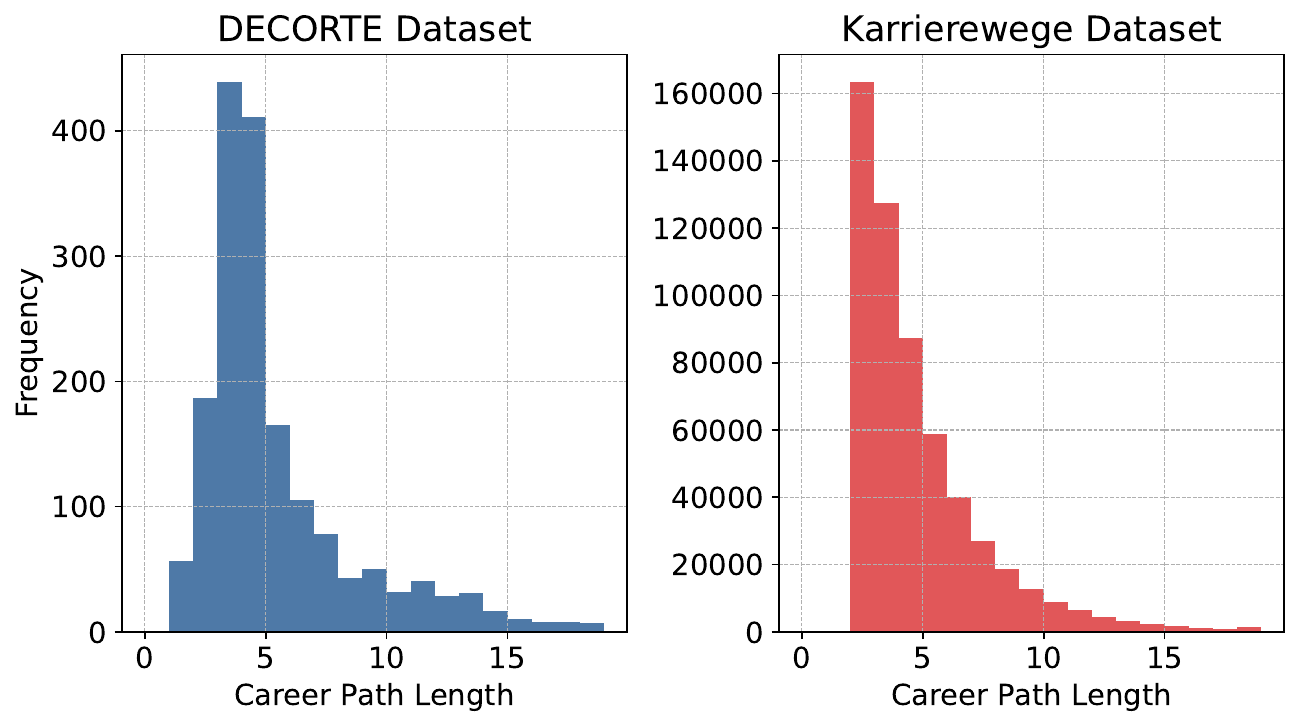}
  \caption{Work experiences per resume for the \textsc{Karrierewege} and DECORTE dataset.}
  \label{fig:jobs_per_resume_our_data}
\end{figure}

\begin{figure}
  \centering
  \includegraphics[width=1.0\linewidth]{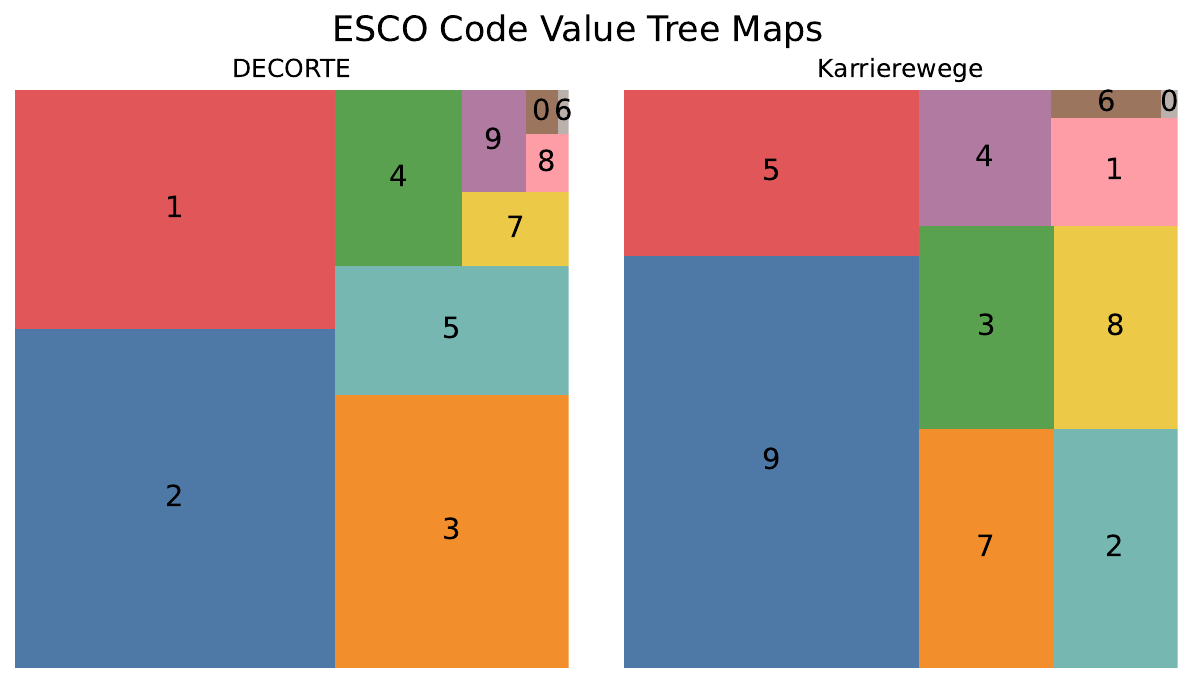}
  \caption{Tree maps on ESCO codes with one digit.}
  \label{fig:datasets_tree_map}
\end{figure}

 \begin{figure}
   \centering
   \includegraphics[width=0.8\linewidth]{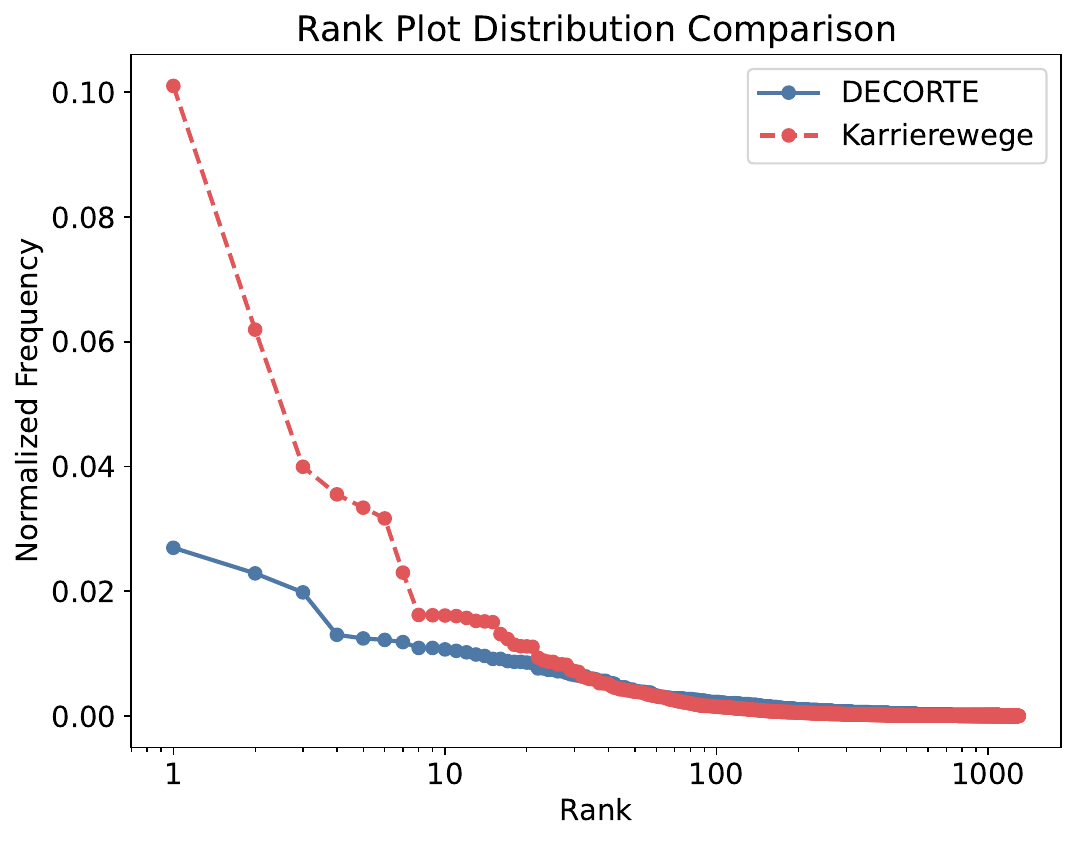}
   \caption{Rank plot of normalized frequencies of ESCO codes with full digits.}
   \label{fig:datasets_rank_plot}
 \end{figure}

On the number and average length of resumes, the \textsc{Karrierewege} dataset (568,888 resumes) contains a higher proportion of resumes with fewer work experiences compared to DECORTE (2,482 resumes), which typically includes five experiences per resume (see Figure \ref{fig:jobs_per_resume_our_data}). 

While on the distribution and diversity of occupations, \textsc{Karrierewege} features 1,295 unique ESCO occupations, whereas DECORTE has 1,102. 
If aggregated by industry using ESCO taxonomy codes, as shown in Figure \ref{fig:datasets_tree_map}, \textsc{Karrierewege}  covers broader economic sectors comprehensively, such as Elementary Occupations (Sector 9) and Service Workers (Sector 5), while DECORTE is concentrated in knowledge workers and managerial activities (Sectors 2 and 1).
Moreover, a rank distribution plot in Figure \ref{fig:datasets_rank_plot} of ESCO occupations further highlights these distinctions: \textsc{Karrierewege}'s most frequent occupations are in Sectors 9, 5, and 8, while DECORTE's are concentrated in Sectors 2 and 1.
Nevertheless, despite these differences in the top occupations, both datasets show similar relative coverage of less frequent occupations, as clearly shown in the tails of both rank distributions. 
When considering absolute frequencies, however, \textsc{Karrierewege} exhibits broader overall sector coverage, even within Sectors 1 and 2. We provide a throughout presentation of these frequencies in Appendix \ref{sec:esco_occupations_concentration} and the full first level ESCO classification names in Appendix \ref{sec:esco_first_level} for completeness.

\begin{figure}
  \centering
  \includegraphics[width=1.0\linewidth]{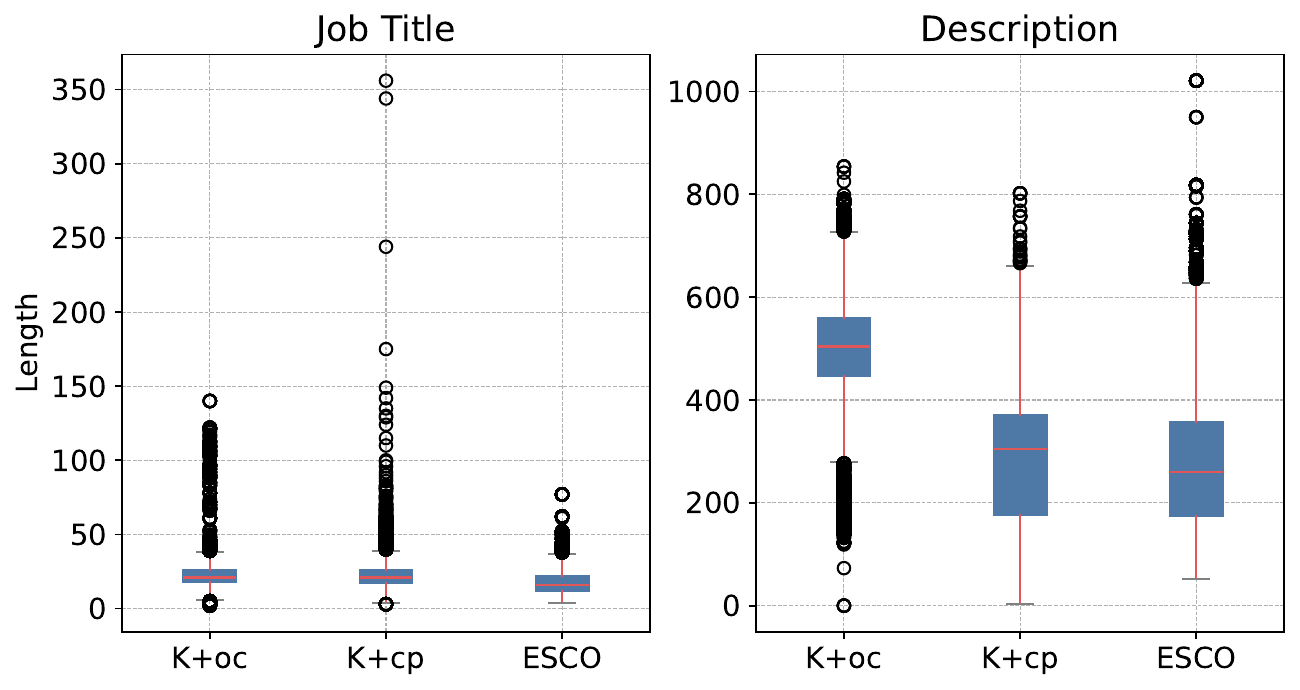}
  \caption{Length of generated job titles and job descriptions with both strategies K+oc and K+cp in comparison with ESCO.}
  \label{fig:datasets_syn_len}
\end{figure}

 
 Regarding the statistics on the synthetic data, Figure \ref{fig:datasets_syn_len} reveals that the lengths of job titles and descriptions differ depending on the data synthetization method, K+oc or K+cp.
 Though differences in job title lengths are minimal, apart from some outliers; K+oc generates consistently longer job descriptions (up to 800 characters), while K+cp produces shorter descriptions, with most under 400 characters. In comparison, ESCO descriptions have comparable length to the K+cp descriptions, while being in general also shorter than the ones generated with method K+oc. 

\section{Benchmark Baseline}
To showcase how well  \textsc{Karrierewege}  supports existing models in realistic career prediction, we adapted a SOTA approach \citep{decorte.2023} to benchmark its performance comprehensively. We want to showcase the effect of the dataset size and the robustness of models trained on \textsc{Karrierewege+} in comparison to the smaller, less diverse DECORTE benchmark dataset.

\subsection{Method}

We follow the scheme called ``LAST'' in \citet{decorte.2023}, and fine-tune a \code{all-mpnet-base-v2}
sentence-transformer model using contrastive representation learning on pairs of text documents: one for the career path $ex_1, \cdots, ex_N$ and another for the ESCO occupation $occ_N$. In \citet{decorte.2023}, each work experience $ex_i$ in a career path is represented as:
\vspace{-.3cm}
\begin{tcolorbox}[colback=gray!20, colframe=white, width=\columnwidth, boxrule=0pt]
role: <title in free-form text> \\
description: <description in free-form text>
\end{tcolorbox}
\vspace{-.3cm}
\hspace{-.4cm}and the career path document is composed by concatenating them with a separator token.
In our adaptation, while the career document is composed similarly, a career experience $ex_i$ is represented as:
\vspace{-.7cm}
\begin{tcolorbox}[colback=gray!20, colframe=white, width=\columnwidth, boxrule=0pt]
esco role: <esco occupation title> \\
description: <esco occupation description>
\end{tcolorbox}
\vspace{-.3cm}
\hspace{-.4cm}In turn, on both approaches, the occupation document is structured always as the latter and contains data from ESCO occupation $occ_N$.

Finally, a linear transformation $T$ is learned by minimizing the least squares error between transformed representations of career paths $ex_1, \cdots, ex_{N-1}$ and representations of their next ESCO occupations $occ_N$. Therefore, a career path prediction over the next occupation is achieved by the scoring function naturally induced by the cosine similarity between a transformed career path representation and all ESCO occupation representations.
While \citet{decorte.2023} include further an ESCO skill overlap component in the scoring function, we opt to leave this out, in order to better measure the impact of only using ESCO data and synthetic free text data in our experiments.
Our full experimental setup is given in Appendix \ref{sec:exp_setup}.

\subsection{Results}
\begin{table*}[t!]
  \centering
\resizebox{2\columnwidth}{!}{
  \begin{tabular}{l|ccc|ccc|ccc}
    \hline
    \textbf{Metric} & \multicolumn{3}{c|}{\textbf{MRR}} & \multicolumn{3}{c|}{\textbf{R@5}} & \multicolumn{3}{c}{\textbf{R@10}} \\
    \hline
    \textbf{Train/ Test} & \textbf{DECORTE} & \textbf{\textsc{K+}oc} & \textbf{\textsc{K+}cp} & \textbf{DECORTE} & \textbf{\textsc{K+}oc} & \textbf{\textsc{K+}cp} & \textbf{DECORTE} & \textbf{\textsc{K+}oc} & \textbf{\textsc{K+}cp} \\
    \hline
    \textbf{DECORTE} & \textbf{0.2427} & 0.1339 & 0.1588 & \textbf{0.3418} & 0.2005 & 0.2302 & \textbf{0.4151} & 0.2669 & 0.3076 \\
    \textbf{\textsc{K+}oc} & \underline{0.1303} & \textbf{0.4312} & \underline{0.3784} & 0.2164 & \textbf{0.5340} & 0.4899 & 0.3091 & \textbf{0.6165} & \underline{0.5784} \\
    \textbf{\textsc{K+}cp} & 0.1294 & \underline{0.3685} & \textbf{0.4281} & \underline{0.2186} & \underline{0.4693} & \textbf{0.5280} & \underline{0.3235} & \underline{0.5566} & \textbf{0.6065} \\
    \hline
  \end{tabular}
  }
  \caption{\label{tab:combined_results}
    Cross evaluation results for MRR, R@5, and R@10 across free-text datasets.
  }
\end{table*}
 To better understand how training data size impacts performance, we experimented with multiple dataset sizes, allowing us to assess trends in model improvement across varying scales. Models trained on \textsc{Karrierewege} consistently achieve higher scores compared to those trained on DECORTE, even when the dataset sizes are identical (see Table \ref{tab:results}). This performance advantage cannot be solely attributed to validation and test set overlap, as the overlap remains negligible, or even minimal for \textsc{Karrierewege+}cp (see Table \ref{tab:overlap} in the Appendix).
This suggests that other patterns in the data, such as the more coherent career paths, might be contributing to the improved results. Across \textsc{Karrierewege}, \textsc{K+}oc, and \textsc{K+}cp, a clear performance improvement is observed with the use of larger training datasets. Notably, the performance increase is most pronounced when scaling from the 2k dataset to the 100k dataset, highlighting the significant impact of additional data. However, once the dataset size reaches a substantial volume, such as 100k, the performance gains taper off, as evidenced in the smaller improvement observed when scaling from 100k to 500k in \textsc{Karrierewege}.
\begin{table}
  \centering
\resizebox{\columnwidth}{!}{
  \begin{tabular}{l|ccc}
    \hline
    \textbf{Dataset} & \textbf{MRR} & \textbf{R@5} & \textbf{R@10} \\
    \hline
    \textbf{DECORTE 2k} & 0.2427 & 0.3418 & 0.4151 \\
    \hdashline
    \textbf{\textsc{K+}oc 2k} & 0.3846 & 0.4779 & 0.5423 \\
    \textbf{\textsc{K+}oc 100k} & \textbf{0.4312} & \textbf{0.5340} & \textbf{0.6165} \\
    \hdashline
    \textbf{\textsc{K+}cp 2k} & 0.3702 & 0.4754 & 0.5568 \\
    \textbf{\textsc{K+}cp 100k} & \textbf{0.4281} & \textbf{0.5280} & \textbf{0.6065} \\
    \hline
    \textbf{DECORTE ESCO 2k} & 0.2084 & 0.2813 & 0.3418 \\
    \hdashline
    \textbf{\textsc{Karrierewege} 2k} & 0.4232 & 0.5146 & 0.5636 \\
    \textbf{\textsc{Karrierewege} 100k} & 0.4775 & 0.5671 & 0.6317 \\
    \textbf{\textsc{Karrierewege} 500k} & \textbf{0.4867} & \textbf{0.5713} & \textbf{0.6347} \\
    \hline
  \end{tabular}
  }
  \caption{\label{tab:results}
    Results for different free-text and standardized resume datasets with their approximate size.
  }
\end{table}

Evaluating across datasets and synthesis methods shows that models trained on \textsc{Karrierewege+} datasets also performed well when tested on DECORTE, indicating that the paraphrased career paths generalize effectively across different datasets (see Table 
\ref{tab:combined_results}
). This strong performance suggests that the paraphrased data captures underlying patterns and relationships between job titles, making it adaptable across various contexts. Notably, models trained on \textsc{K+}cp datasets generalize better than models trained on \textsc{K+}oc, further supporting the idea that coherent career paths play a role in improving model performance. 

\section{Conclusions and Further Research}
We introduced \textsc{Karrierewege} and \textsc{Karrierewege+}, large-scale, publicly available datasets for career path prediction. By linking the datasets to the ESCO taxonomy and synthesizing paraphrased job titles and descriptions, we addressed the challenge of predicting career trajectories from the free-text inputs typically found in resumes. Our results demonstrate that models trained on the \textsc{Karrierewege} datasets, particularly the \textsc{Karrierewege+}cp variant, perform well on free-text data, underscoring the importance of data diversity and richness for accurate career path prediction.

While these datasets provide a strong foundation for model training and evaluation, future work could focus on expanding their scope to include more regions, and languages, further enhancing their applicability to global career path prediction. Addressing challenges such as cross-industry career transitions could also improve model robustness and generalizability.


\FloatBarrier
\section*{Ethical Considerations}

The use of large-scale datasets like \textsc{Karrierewege} carries the risk of bias amplification if the dataset overrepresents certain industries, job levels, or demographics. Biases in the dataset could inadvertently lead to discriminatory predictions, particularly when applied to automated decision-making tools used by recruiters or employment agencies. Furthermore, synthesizing data to augment or enhance the dataset introduces additional risks, as the assumptions made by the underlying models may reflect or amplify existing biases. This could result in inaccurate or skewed descriptions of career trajectories, reinforcing stereotypes or marginalizing underrepresented groups.

To address these challenges, it is crucial to continuously monitor and mitigate biases that may emerge both in the original dataset and in any synthesized data. Strategies such as bias audits, fairness metrics, and diversification of training data sources should be implemented to ensure equitable model predictions.


\section*{Acknowledgements}
We thank the reviewers for their insightful feedback. We thank Jan-Peter Bergmann for computing infrastructure support and advice. ES and YC acknowledges financial support with funds provided by the German Federal Ministry for Economic Affairs and Climate Action due to an enactment of the German Bundestag under grant 46SKD127X (GENESIS). BP is supported by ERC Consolidator Grant DIALECT 101043235.



\bibliography{custom}

\appendix

\section{Prompts for linkage}
\label{sec:prompts-linkage}
To address the limited context window of the GPT models, we used ISCO-08 codes as a filtering mechanism to narrow down potential occupation matches between Berufenet and ESCO. By applying the ISCO code, the number of candidate pairs was significantly reduced, allowing them to fit within the GPT models' context window. In some cases, however, better matches were found under different ISCO codes. Therefore, ISCO filtering was only applied when necessary to reduce the number of matches. Without this, each Berufenet occupation could potentially match up to 3,039 ESCO occupations.

\subsection{English Prompt}

\begin{lstlisting}
###CONTEXT###
You are a specialist in matching occupations with their ESCO labels. I have tried 3 different methods, and each method has a different prediction for the label. You will be presented with the occupation title, the 3 different predictions, and the set of candidate labels. The language is German.

###INSTRUCTION###
Choose the most appropriate label for an occupation title.
Return a JSON object with the job title and the selected label.

###DATA###
occupation_title: {occupation}
pred_1: {pred_1}
pred_2: {pred_2}
pred_3: {pred_3}
all_candidates: {candidates}

###OUTPUT###
\end{lstlisting}

\subsection{German Prompt}

\begin{lstlisting}
###CONTEXT###
Du bist ein Spezialist für das Matching von Berufen mit ihren ESCO-Labels. Ich habe 3 verschiedene Methoden ausprobiert, und jede Methode hat eine andere Vorhersage für das Label. Dir werden die Berufsbezeichnung, die 3 verschiedenen Vorhersagen und die Menge der in Frage kommenden Bezeichnungen vorgelegt. Die Sprache ist gemischt zwischen Deutsch.

###ANLEITUNG###
Wähle das am besten geeignete Label für eine Berufsbezeichnung.
Gib ein JSON-Objekt mit dem Berufstitel und dem gewählten Label zurück.

###DATA###
beruf_title: {occupation}
pred_1: {pred_1}
pred_2: {pred_2}
pred_3: {pred_3}
all_candidates: {candidates}

###OUTPUT###
\end{lstlisting}

\section{Prompts for title and description generation}
\label{sec:prompts-generation}
\subsection{Generation per career path}
\begin{lstlisting}
"""
    Please create a paraphrased version of the following career path: {job_list}.
    The length of the list should be the same as the original list. Please adhere to the format and do not add anything else.

    ###
    Format:

    'Career Path:

    [paraphrased title 1, paraphrased title 2, ...]'


    """
\end{lstlisting}
\begin{lstlisting}
"""
    Please create a description for the following jobs: {job_list}.
    Please make it not longer than 4 sentences. Please adhere to the format and do not add anything else.

    ###
    Format:

    'Descriptions: 

    [Job title 1: Description of job 1,
    
    Job title 2: Description of job 2,
    
    ...]'
    """
\end{lstlisting}
\subsection{Generation per occupation}
\begin{lstlisting}
"""
    Paraphrase the following occupation title: {job}.
    Please return a list of max 7 alternative job titles.
    ###
    Example:
    occupation title: physiotherapist

    1. Physical Therapist
    2. Physiotherapy Specialist
    3. Rehabilitation Therapist
    4. Movement Therapist
    5. Injury Recovery Specialist
    6. Musculoskeletal Therapist
    7. Sports Medicine Therapist
    """
\end{lstlisting}
\begin{lstlisting}
"""
    Please create a description for the following job: {job}.
    Please make it not longer than 4 sentences. Please adhere to the format and do not add anything else.

    ###
    Format:

    'Description: 

    Job Description of the occupation...'
    """
\end{lstlisting}

\section{Prompts for LLM-evaluation}
\label{sec:prompts-evaluation}

\subsection{Prompt for all metrics at once}
\begin{lstlisting}
"""
    # CONTEXT
    A paraphrased career path should accurately reflect the skills and tasks of the original career path and it's job titles. 
    You will evaluate the paraphrased career path on the following four dimensions: 
    Correctness measures if the paraphrased titles are accurate representations of the original job titles. 
    A paraphrased career path with high correctness means that all titles are job titles that
    exist in reality, but aren't just copies of 
    the original titles. A low-correctness paraphrased career path may contain titles that are not job titles or are the same title as the original title.
    Semantic similarity assesses how well the paraphrased job titles captures the meaning of the original job titles in a career path. 
    A high semantic similarity score means that the paraphrased titles accurately represent the skills and tasks of the original titles.
    Diversity measures how many unique job titles are present in the paraphrased career path. 
    A high diversity score means that the paraphrased career path contains a wide range of job titles and does not contain the same title multiple times.
    Coherence evaluates how well the paraphrased job titles fit together within the paraphrased career path. 
    A highly coherent paraphrased career path will have job titles that make sense together or form a logical progression.
.

    # INSTRUCTIONS
    You will be presented with a original career path and its paraphrased version.
    For each dimension, give the summary a score between 0 and 5. For correctness a score of 0 means 
    the paraphrased career path contains only job titles that aren't real job titles or just copies of the original title, while a score of 5 means it 
    provides only correct job titles.
    For semantic similarity a score of 0 means that all paraphrased job titles are do not capture the meaning of the original titles very well, while a score of 5 means that all paraphrased job titles accurately represent the skills and tasks of the original titles. 
    For Diversity a score of 0 means that the paraphrased career path contains the same title multiple times, while a score of 5 means that the paraphrased career path contains a wide range of job titles.
    For Coherence a score of 0 means that the paraphrased job titles do not make sense together or form a logical progression, while a score of 5 means that the paraphrased job titles fit together well in the career path.
    Output the Likert scores for each dimension as a json (key: dimension, value: likert-score).
    Do not add any explanation, answer only the Likert scores.
    Use the initial marker ```json and the final marker ``` to mark the json content.\n\n

    # EVALUATION MATERIALS
    ## Original career path
    {original_career_path\}

    ## paraphrased career path
    {paraphrased_career_path\}
    """
\end{lstlisting}

\subsection{Prompts for each metric}
\begin{lstlisting}
"""
    # CONTEXT
    Correctness measures if the paraphrased titles are accurate representations of the original job titles. 
    A paraphrased career path with high correctness means that all titles are job titles that
    exist in reality, but aren't just copies of 
    the original titles. A low-correctness paraphrased career path may contain titles that are not job titles or are the same title as the original title.

    # INSTRUCTIONS
    You will be presented with a original career path and its paraphrased version.
    Give the summary a score between 0 and 5. 
    Zero means the paraphrased career path contains only job titles that aren't real job titles or just copies of the original title, 
    while a score of 5 means it provides only correct job titles.
    Just answer with the Likert Score, no text please.

    # EVALUATION MATERIALS
    ## Original career path
    {original_career_path}

    ## paraphrased career path
    {paraphrased_career_path}
    """
\end{lstlisting}
\begin{lstlisting}
"""
    # CONTEXT
    Semantic similarity assesses how well the paraphrased job titles captures the meaning of the original job titles in a career path. 
    A high semantic similarity score means that the paraphrased titles accurately represent the skills and tasks of the original titles.

    # INSTRUCTIONS
    You will be presented with a original career path and its paraphrased version.
    Give the summary a score between 0 and 5. 
    Zero means that all paraphrased job titles are do not capture the meaning of the original titles very well, 
    while a score of 5 means that all paraphrased job titles accurately represent the skills and tasks of the original titles.
    Just answer with the Likert Score, no text please.

    # EVALUATION MATERIALS
    ## Original career path
    {original_career_path}

    ## paraphrased career path
    {paraphrased_career_path}
    """
\end{lstlisting}
\begin{lstlisting}
"""
    # CONTEXT
    Diversity measures how many unique job titles are present in the paraphrased career path. 
    A high diversity score means that the paraphrased career path contains a wide range of job titles and does not contain the same title multiple times.

    # INSTRUCTIONS
    You will be presented with a paraphrased career path.
    Give the summary a score between 0 and 5. 
    Zero means that the paraphrased career path contains the same title multiple times, 
    while a score of 5 means that the paraphrased career path contains a wide range of job titles.
    Just answer with the Likert Score, no text please.

    ## paraphrased career path
    {paraphrased_career_path}
    """
\end{lstlisting}
\begin{lstlisting}
"""
    # CONTEXT
    Coherence evaluates how well the paraphrased job titles fit together in the career path. 
    A highly coherent paraphrased career path will have job titles that make sense together or form a logical progression.

    # INSTRUCTIONS
    You will be presented with a paraphrased career path.
    Give the summary a score between 0 and 5. 
    Zero means that the paraphrased job titles do not make sense together or form a logical progression, 
    while a score of 5 means that the paraphrased job titles fit together well in the career path.
    Just answer with the Likert Score, no text please.

    ## paraphrased career path
    {paraphrased_career_path}
    """
\end{lstlisting}
\section{Alignment Scores}
\label{sec:appendix_alignement_scores}
Table \ref{tab:llm-judge} shows the Likert scale score for the human labeling as well as the \code{gpt-4o-mini} results. Table \ref{tab:kappa} compares Kappa scores between \code{gpt-4o-mini} with one prompt and \code{gpt-4o-mini} with one prompt per metric. Table \ref{tab:rouge_bleu} presents BLEU and ROUGE-L scores for job titles and descriptions.

\begin{table}[h!]
\resizebox{\columnwidth}{!}{
\begin{tabular}{l|c|c}
\hline
\textbf{Metric} & \textbf{gpt-4o-mini one Prompt} & \textbf{gpt-4o-mini} \\ \hline
Correctness (CP) & 0.367089 & 0.058639 \\ \hline
Semantic Similarity (CP) & 0.116162 & -0.047056 \\ \hline
Diversity (CP) & 0.312162 & 0.264043 \\ \hline
Coherence (CP) & 0.082716 & -0.013099 \\ \hline
Correctness (Free) & 0.082840 & 0.100846 \\ \hline
Semantic Similarity (Free) & 0.022131 & -0.047251 \\ \hline
Diversity (Free) & 0.389831 & 0.333567 \\ \hline
Coherence (Free) & 0.022483 & -0.025326 \\ \hline
\end{tabular}
}
\caption{\label{tab:kappa}Comparison of Kappa scores between LLM with one prompt and LLM with one prompt per metric.}
\end{table}

\begin{table}[h!]
\resizebox{\columnwidth}{!}{
\begin{tabular}{l|c|c}
\hline
\textbf{Metric}                       & \textbf{\textsc{Karrierewege+}cp} & \textbf{\textsc{Karrierewege+}oc} \\ \hline
ROUGE-L Score (Job Titles)    & 0.1618      & 0.1592      \\ \hline
ROUGE-L Score (Job Descriptions) & 0.0426   & 0.0233      \\ \hline
BLEU Score (Job Titles)       & 0.0005      & 0.0002      \\ \hline
BLEU Score (Job Descriptions) & 0.0005      & 0.0003 
\\ \hline
\end{tabular}
}
\caption{\label{tab:rouge_bleu}BLEU and ROUGE-L scores for job titles and descriptions.}
\end{table}
\begin{table*}
\resizebox{2\columnwidth}{!}{
\begin{tabular}{l|c|c|c|c}
\hline
\textbf{Metric} & \textbf{Manual Labeling (100)} & \textbf{gpt-4o-mini (100)} & \textbf{gpt-4o-mini one Prompt (100)}  & \textbf{gpt-4o-mini one Prompt (all)}\\ \hline
Mean Correctness (CP) & 4.82 & 4.31 & 4.72 & 4.72 \\ \hline
Mean Correctness (OCC) & 4.72 & 4.26 & 4.74 & 4.70\\ \hline
Mean Semantic Similarity (CP) & 4.50 & 3.41 & 4.02 & 4.08 \\ \hline
Mean Semantic Similarity (OCC) & 4.20 & 2.91 & 3.91 & 3.85 \\ \hline
Mean Diversity (CP) & 4.08 & 3.38 & 4.01 & 3.85\\ \hline
Mean Diversity (OCC) & 4.43 & 3.88 & 4.33 & 4.22 \\ \hline
Mean Coherence (CP) & 4.01 & 1.58 & 4.04 & 4.12 \\ \hline
Mean Coherence (OCC) & 3.95 & 1.44 & 3.95 & 3.93 \\ \hline
\end{tabular}
}
\caption{\label{tab:llm-judge}Comparison of Mean Scores between Manual Labeling, the LLM-as-a-judge with one prompt per metric (gpt-4o-mini) and the LLM-as-a-judge with one one prompt for all metrics (gpt-4o-mini one Prompt).}
\end{table*}

\section{Experimental Setup}
\label{sec:exp_setup}

Following the recipe in \cite{decorte.2023}, we fine tune the model using Multiple Negatives Ranking Loss (MNRL), in-batch negatives and augmented career path data with all possible sub-paths of minimum length 2. This augmentation is applied after the data split. The fine-tuning process is conducted using a batch size of $16$, a learning rate of $2e-5$, for up to $1$ epoch for the large and $2$ epochs for the small datasets, with evaluation every $1$\% of steps based on validation loss. The best-performing model is saved based on these evaluations.

\section{Overlap Test and Validation Dataset}
\label{sec:overlap}
Table \ref{tab:overlap} shows the overlap between validation and test splits for various datasets.

\begin{table*}[h!]
  \centering
  \begin{tabular}{l|c|c|c|c}
    \hline
    \textbf{Dataset} & \textbf{Length Validation} & \textbf{Length Test} & \textbf{Overlap} & \textbf{Overlap \%} \\
    \hline
    DECORTE ESCO & 1,558 & 1,801 & 45 & 2.92\% \\
    \hline
    DECORTE & 1,558 & 1,801 & 0 & 0\% \\
    \hline
    \textsc{Karrierewege}  & 667,404 & 658,012 & 56,101 & 8.53\% \\
    \hline
    \textsc{Karrierewege+}oc & 138,275 & 137,530 & 8,724 & 6.34\% \\
    \hline
    \textsc{Karrierewege+}cp &  138,275 & 137,530 & 139 & 0.10\% \\
    \hline
  \end{tabular}
  \caption{\label{tab:overlap}
    Overlap between validation and test splits across various datasets. The percentage is calculated as the number of overlapping entries divided by the total size of the test split.
  }
\end{table*}


\section{First level of the ESCO classification}
\label{sec:esco_first_level}
Table \ref{tab:esco_first_level} shows the first categorization level of the ESCO classification with their respective codes.

\begin{table}[h!]
\centering
\begin{tabular}{l|l}
\hline
\textbf{Code} & \textbf{First Level Occupation Category} \\ 
\hline
0 & Armed forces occupations \\
1 & Managers \\
2 & Professionals \\
3 & Technicians and associate \\
  & professionals \\
4 & Clerical support workers \\
5 & Service and sales workers \\
6 & Skilled agricultural, forestry \\
  & and fishery workers \\
7 & Craft and related trades workers \\
8 & Plant and machine operators \\
  & and assemblers \\
9 & Elementary occupations \\
\hline
\end{tabular}
\caption{First level of the ESCO classification.}
\label{tab:esco_first_level}
\end{table}

\section{Absolute statistics of ESCO occupations distribution}
\label{sec:esco_occupations_concentration}


As presented in Tables \ref{tab:karrierewege_level_1_dist} and \ref{tab:decorte_level_1_dist},
Sectors 1 and 2 have almost two order of magnitude higher absolute frequencies in  \textsc{Karrierewege} when compared to DECORTE data.
By observing their absolute numbers, the massive difference between both datasets become clearer and highlights the potential of \textsc{Karrierewege}.

\begin{table}[h!]
\centering
\begin{tabular}{l|r}
\hline
\textbf{Code} & \vtop{\hbox{\strut \textbf{Absolute}}\hbox{\strut \textbf{Frequency}}} \\ 
\hline
9 & 941544 \\
5 & 379332 \\
7 & 247981 \\
2 & 230291 \\
3 & 213148 \\
8 & 194963 \\
4 & 139843 \\
1 & 106259 \\
6 &  23574 \\
0 &   3434 \\
\hline
\end{tabular}
\caption{Absolute frequency of first level ESCO occupations in \textsc{Karrierewege}.}
\label{tab:karrierewege_level_1_dist}
\end{table}

\begin{table}[h!]
\centering
\begin{tabular}{l|r}
\hline
\textbf{Code} & \vtop{\hbox{\strut \textbf{Absolute}}\hbox{\strut \textbf{Frequency}}} \\ 
\hline
2  & 2891 \\
1  & 2036 \\
3  & 1699 \\
5  &  807 \\
4  &  600 \\
7  &  210 \\
9  &  172 \\
8  &   66 \\
0  &   39 \\
6  &   12 \\
\hline
\end{tabular}
\caption{Absolute frequency of first level ESCO occupations in DECORTE.}
\label{tab:decorte_level_1_dist}
\end{table}

\end{document}